\pdfoutput=1

\documentclass[11pt,a4paper,dvipsnames,table]{article}

\usepackage[]{acl}
\usepackage[T1]{fontenc}
\usepackage[utf8]{inputenc}
\usepackage{microtype}

\usepackage{times}
\usepackage{bm}
\usepackage{latexsym}
\usepackage{graphicx}
\usepackage{amsmath}
\usepackage{multirow}
\usepackage{xcolor}
\usepackage{dsfont}
\usepackage{dingbat}
\usepackage{amssymb}
\usepackage{booktabs}
\usepackage{cleveref}
\usepackage{makecell}
\usepackage{lipsum,multicol}
\usepackage{mathtools}

\usepackage[english]{babel}
\usepackage{hyperref}
\addto\extrasenglish{

}
\usepackage{enumitem,kantlipsum}

\usepackage{todonotes} 

\usepackage{microtype}
\usepackage{bbm}
\usepackage[english]{babel}
\usepackage{amsthm}
\newtheorem{proposition}{Proposition}
\theoremstyle{remark}

\newtheorem{assumption}{Assumption}

\newtheorem{axiom}{Axiom}

\title{Logical Satisfiability of Counterfactuals for Faithful Explanations in NLI}

\author{
    Suzanna Sia $^1$\thanks{\; Work done while at internship at Meta AI Research, corresponding email ssia1@jhu.edu.} \qquad Anton Belyy $^1$ \qquad Amjad Almahairi $^2$ \\
    \bf Madian Khabsa $^2$ \qquad Luke Zettlemoyer $^2$ \qquad Lambert Mathias $^2$  \\
    \\[.1em]
    $^1$Johns Hopkins University \;\;
    $^2$Meta AI Research \;\;
}

\date{}

\begin{document}
\maketitle
\begin{abstract}
Evaluating an explanation's faithfulness is desired for many reasons such as trust, interpretability and diagnosing the sources of model's errors. 
In this work, which focuses on the NLI task, we introduce the methodology of Faithfulness-through-Counterfactuals, which first generates a counterfactual hypothesis based on the logical predicates expressed in the explanation, and then evaluates if the model's prediction on the counterfactual is consistent with that expressed logic (i.e. if the new formula is \textit{logically satisfiable}). In contrast to existing approaches, this does not require any explanations for training a separate verification model. We first validate the efficacy of automatic counterfactual hypothesis generation, leveraging on the few-shot priming paradigm. Next, we show that our proposed metric distinguishes between human-model agreement and disagreement on new counterfactual input. In addition, we conduct a sensitivity analysis to validate that our metric is sensitive to unfaithful explanations.  
\end{abstract}

\section{Introduction}
\label{sec:introduction}

How should we evaluate an explanation's \textit{faithfulness} with respect to the task model? According to \citet{jacovi2020towards}, faithful measures should focus on utility to the user and the idea that an explanation can be \textit{sufficiently faithful}.\footnote{\citet{jacovi2020towards} originally posit that faithful explanations should ``accurately represents the reasoning process behind the model’s prediction'', however also acknowledge that this is ``impossible to satisfy fully''.} The goal of interpretability research is to increase \textit{warranted} trust and identify the influence of certain variables and allow users to understand how a model will behave on given inputs  \cite{doshi2017towards, lipton2018mythos}.  

In interpretable NLP,  there is growing interest in tasks that require world and commonsense ``knowledge'' and ``reasoning'' \cite{danilevsky2020survey}. We focus on natural language inference (SNLI; \citet{bowman2015large}), where extractive explanations also known as rationales \cite{deyoung2019eraser} are limited as they take a subset of the existing input. Instead, we require \textit{free-form natural language explanations} to fill in the reasoning or knowledge gap for such tasks \cite{camburu2018snli,rajani2019explain}. Our setting is thus characterised by the post-hoc interpretation of black-box classification models via generative explanations. Our work follows the standard “predict-and-explain” paradigm \cite{do2020snli}. Here an explanation generator generates the explanations conditioned on the predicted task label.\footnote{\newcite{do2020snli} report little to no difference between jointly predict and explain compared to predict then explain. Note that the emphasis of this work is on evaluating the faithfulness of explanations rather than generating them.} Without faithfulness evaluations, the explanation approximately describes the internal process at best, and is generated from superficial similarities between the training data and the class label at worst.\footnote{Preliminary experiments show that flipping the input class label can change the form of the explanation.}

\begin{figure}[t]
    \centering
    \includegraphics[width=0.485\textwidth]{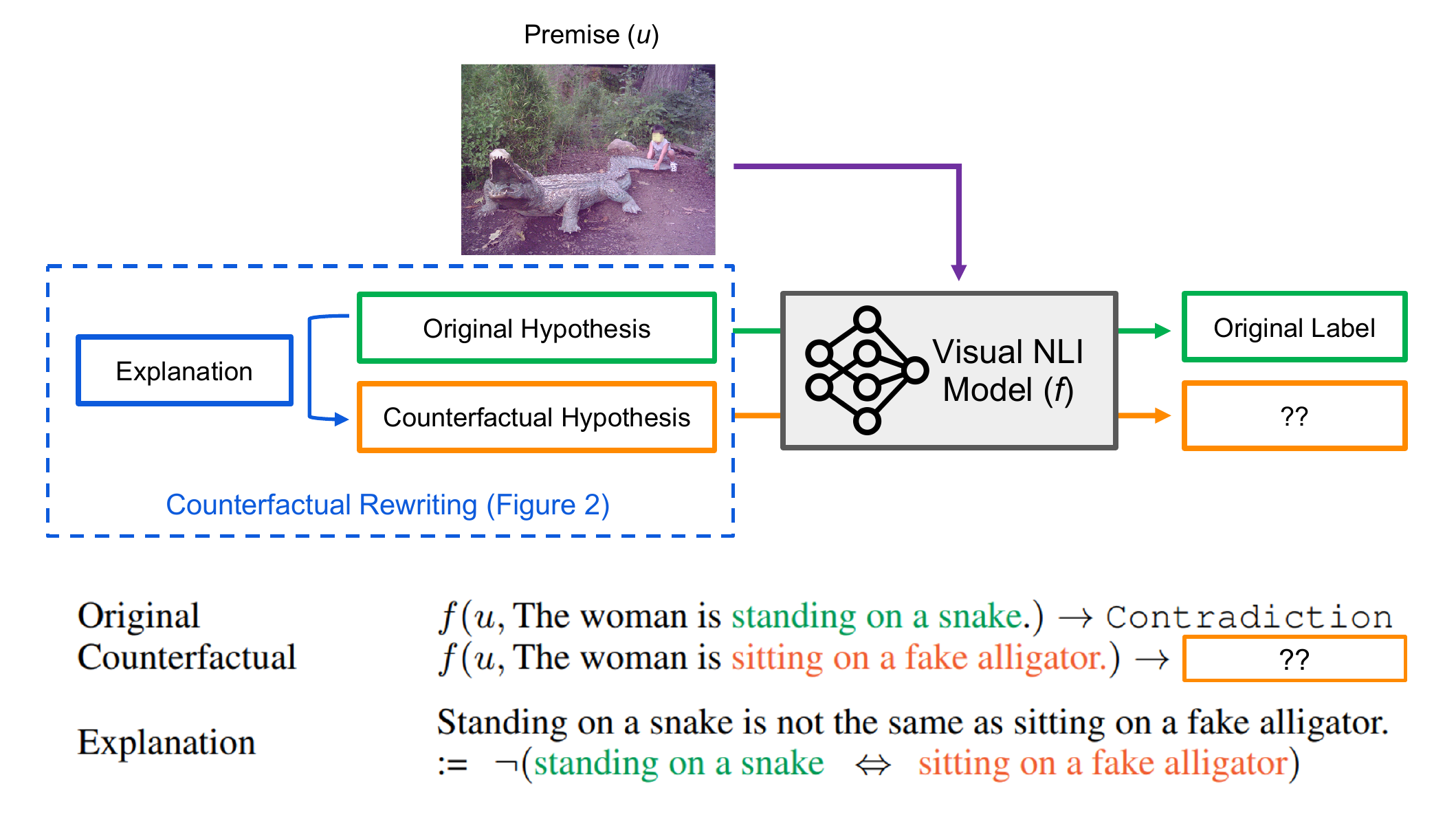}
    \caption{\small Overview of the proposed FTC approach, evaluating \textbf{f}aithfulness of explanations \textbf{t}hrough \textbf{c}ounterfactuals. 
    If the explanation is faithful to the model, the NLI label on the new counterfactual hypothesis should change to \texttt{Entailment}. If the model still predicts \texttt{Contradiction}, this indicates that the explanation is not faithful to the model, i.e. the logic of the explanation and the model are not consistent.
    }
    \label{fig:fig1}
\end{figure}

The central contribution of this paper is a \textbf{ methodology grounded in first-order free logic  \cite{lambert1967free, bencivenga2002free},\footnote{This is an extension of predicate logic and should not be confused with either predicate or propositional logic.} to verify a given explanations’ faithfulness}.  
Our proposed approach generates a revised (counterfactual) hypothesis based on the logical propositions expressed in the explanation, and evaluates the \textit{logical satisfiability} \cite{boolos2002computability} of the new hypothesis. Consider the following example: 

\vspace{0.5em}
\resizebox{0.95\columnwidth}{!}{
\begin{tabular}{ll}
    \toprule
    \textbf{Hypothesis:} & The dog is barking at the girl.\\
    \textbf{Explanation:} & The dog is an animal.\\
    \textbf{Counterfactual:} & The animal is barking at the girl.\\
    \bottomrule
\end{tabular}
}
\vspace{0.5em}


If the explanation is logically consistent (faithful) to the model, then the revised counterfactual hypothesis which replaces `dog' with `animal' in the original hypothesis, should be satisfiable, since `dog is an animal'. However, if the explanation is inconsistent with the model, then the resulting hypothesis is unsatisfiable and the explanation is unfaithful. We describe this formally in \autoref{sec:method}.

Compared to previous automatic metrics (LAS; \citet{hase2020leakage}, LRA; \citet{wiegreffe2020measuring}), our proposed method does not rely on an external verification model and therefore does not require explanation data for training.\footnote{``Faithfulness'' measures which are tied to an external verification model are potentially problematic as given a fixed task model and explanation, one could in theory achieve two different faithfulness scores if the verification model changes.} Our method directly queries the task model in question while crucially avoiding the confound of ``label leakage''\footnote{Label leakage occurs because of superficial similarities between the syntatic form of the explanation and the task label. For instance the explanations ``A is a B'' and ``A is not a B'' are highly associated with the Entailment and Contradiction label.} from the explanation \cite{hase2020leakage}. We expand on this discussion in \autoref{sec:related_work}. 
The contributions of this work are as follows:

\begin{itemize}
    \item We present a methodology for evaluating faithfulness of free form explanations for NLI, grounded in first-order free logic (\autoref{sec:problem_formulation}, \autoref{sec:satisfiability}). Our method evaluates the satisfiability of logical relations expressed through a counterfactual hypothesis. 

    \item We introduce an automatic metric (\autoref{sec:metric}) and show its viability with human studies, indicating that a practical solution exists for the proposed (theoretical) method. We leverage few-shot priming for generating counterfactual hypothesis, achieving $0.71-0.88$ METEOR score for human and generated explanations (\autoref{sec:validation_span}). 
    
    \item We further evaluate the effectiveness of the fully automated approach for the proposed metric in distinguishing when the model's prediction agrees and disagrees with gold labels on gold counterfactual hypothesis, and achieve $0.58-0.75$ $\rho$-statistic on Wilcoxon rank-sum test (\autoref{sec:validation_entailment}). A further sensitivity analysis indicates that our method is sensitive to pathological explanations that were generated by removing inputs to the explanation generator, as compared to other existing faithfulness metrics (\autoref{sec:sensitivity_analysis}). 
\end{itemize}

\section{Method}
\label{sec:method}

\begin{table}[!t]
\centering
\small
\resizebox{\columnwidth}{!}{ 
\begin{tabular}{ p{0.1\linewidth}p{0.25\linewidth}p{0.5\linewidth}} 
 Label & Propositions & Description \\
 \toprule
\texttt{E} & $u \Rightarrow x$ & hypothesis implied by premise \\

\texttt{C} & $u \Rightarrow \neg x$ &  hypothesis contradicts premise\\
 \texttt{N} & $u \xRightarrow{\textbf{(?)}} x$ &  hypothesis neither contradicts or is entailed by premise \\
\end{tabular}}
\caption{Mapping NLI task labels to propositions. $\Rightarrow$ indicates logical implication, $\neg$ indicates logical negation, and $\xRightarrow{\textbf{(?)}}$ indicates \textit{truth-valueless}.}
\label{tab:labelmap}
\end{table}

\subsection{Problem Formulation}
\label{sec:problem_formulation}

\begin{table*}[!t]
\centering
\small
\begin{tabular}{ll@{\hskip 0.5in}ll@{\hskip 0.5in}ll} 
 \toprule
  Label ($y$) & \makecell[l]{Propositional \\ Formula} & Explanation ($z$) & $R(A, B)$ & \makecell[l]{Propositional \\ Formula (cf) } & Label ($y^{\textrm{cf}}$) \\ 
\midrule
\texttt{E} & $u \Rightarrow x$ & $A$ is the same as $B$ & $A \Leftrightarrow B$ &  $u \Rightarrow x^{\mathrm{cf}}$ & \texttt{E} \\ 
\texttt{C} & $u \Rightarrow \neg x$ & $A$ is not $B$ & $\neg(A \Leftrightarrow B)$ & $u \Rightarrow \neg(\neg x^{\mathrm{cf}})$ & \texttt{E} \\ 
 \multirow{2}{*}{\texttt{N}} & \multirow{2}{*}{ $u \xRightarrow{\textbf{(?)}} x$} & \multirow{2}{*}{$A$ does not imply $B$} & \multirow{2}{*}{\makecell[l]{$\neg(A \Rightarrow B)$}}
 & $u \Rightarrow  x^{\mathrm{cf}}_{[A]}$ & \texttt{E} \\ 
 &  &  & &  $u \xRightarrow{\textbf{(?)}} x^{\textrm{cf}}_{[B]}$ & \texttt{N} \\ 

 \bottomrule
\end{tabular}

\caption{From the original hypothesis $x$ and logical predicate relations $R(A,B)$ expressed in the explanation, we generate the counterfactual hypothesis $x^{\textrm{cf}}$ (2 cases $x^{\textrm{cf}}_{[A]}$ and $x^{\textrm{cf}}_{[B]}$ for $y=\texttt{N}$). The resulting counterfactual label $y^{\textrm{cf}}$ is logically derived from the propositions \autoref{sec:satisfiability}. \autoref{tab:examples} shows examples of this process.}

\label{tab:cf-hypothesis}
\end{table*}

Natural Language Inference (NLI) is typically cast as a classification task; given a premise $u$ and a hypothesis $x$, the classifier $f$ predicts the label $y$, where $y \in \{\texttt{E}, \texttt{C}, \texttt{N}\}$. Here \texttt{E} indicates entailment,  \texttt{C} contradiction, and \texttt{N} neutral, for the relationship between $u$ and $x$. $f$ can therefore be viewed as a black-box function approximating the solution to a \textit{logical satisfiability} problem. 

\paragraph{Testing Predicate Relations in Explanations.}

An explanation $z$ can express one or more logical predicate relations ($R$), which describes the relationship between two variables $A$ and $B$ that are expressed in $x$ and $u$ respectively (see \autoref{tab:cf-hypothesis} col 4 for examples). This is denoted as $R(A, B)$.

A ``faithful'' explanation with respect to a task model $f$, is one that expresses predicate relations $R(A,B)$ that are \textit{consistent} with $f$'s predictions. 
The central idea of this work, is to automatically verify this using a \textit{counterfactual hypothesis}, $x^{\textrm{cf}}$ and its derived associated \textit{counterfactual label} (the expected satisfiability result). If $f(u, x^{\textrm{cf}})$ does not result in the associated counterfactual label, then the explanation is not faithful to $f$. 

\paragraph{First-order Free Logic in NLI.}

In order to deduce the associated label for $x^{\textrm{cf}}$, we must address the issue of logical deduction for \texttt{Neutral}, which has no corresponding expression in classical predicate logic. 
We observe that the ternary label in NLI parallels first-order free logic, which has three distinct logical forms, positive, negative and neutral \cite{lambert1967free,  sep-logic-free}. In contrast to classical logic which requires each singular term to denote a Boolean variable in the domain, free logic may have formula which are \textit{truth-valueless} \cite{sep-logic-free}, i.e., it is not known whether they are True or False.\footnote{Truth-valueless formulas are often said to have ``truth-value gaps''. Informally, this can be interpreted as there being insufficient information on the truth-values of logical variables to conclude the relationship between $u$ and $x$ \cite{sep-logic-free}.} \autoref{tab:labelmap} shows the task label, propositions and their meaning in Free Logic.



\begin{table*}[!t]
\small
\centering
\begin{tabular}{llp{10cm}}

 


\multirow{3}{*}[-2px]{\includegraphics[width=75px]{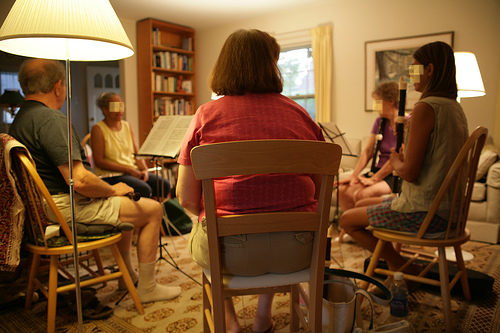}} & Original  &  $f(u, \textrm{There are people \textcolor{ForestGreen}{playing the piano}}) \rightarrow \texttt{Contradict}$ \\
& Counterfactual  &  $f(u, \textrm{There are people \textcolor{RedOrange}{playing woodwind instruments.}})\rightarrow \texttt{Entail}$\\

\noalign{\vskip 2mm}    
& \multirow{2}{*}{Explanation} & \textit{``The people in the photo are not playing the piano. They are instead playing other woodwind instruments.''} \\ 
& & := \hspace{0.2em} $\neg(\textrm{\textcolor{ForestGreen}{playing the piano}}$ \hspace{0.3em} $\Leftrightarrow$ \hspace{0.2em}  $\textrm{\textcolor{RedOrange}{playing other woodwind instruments}})$ \\
\midrule

\rule{0pt}{1.5em}\multirow{3}{*}[0.5em]{\includegraphics[width=75px]{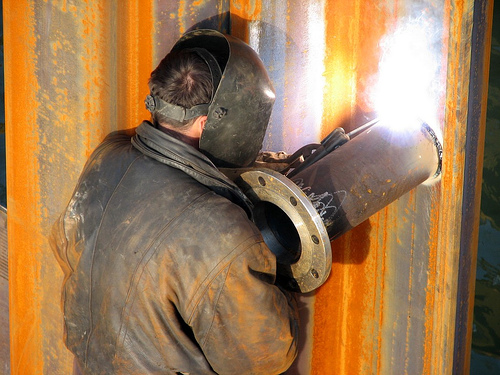}} & Original &  $f(u, \textrm{A man in \textcolor{ForestGreen}{safety equipment} is working on a piece of metal.})\rightarrow \texttt{Entail}$ \\
& Counterfactual &  $ f(u, \textrm{A man in \textcolor{RedOrange}{protective gear} is working on a piece of metal.})\rightarrow  \texttt{Entail}$ \\
\noalign{\vskip 2mm}    
& \multirow{2}{*}{Explanation} & \textit{``Protective gear is a piece of safety equipment''}  \\ 
& & := \hspace{0.2em} $(\textrm{\textcolor{ForestGreen}{safety equipment}}$ \hspace{0.3em} $\Leftrightarrow$ \hspace{.2em}  $\textrm{\textcolor{RedOrange}{protective gear}})$ \vspace{0.8em} \\

\midrule

\rule{0pt}{1.1em}\multirow{3}{*}[0em]{\includegraphics[width=75px,height=60px]{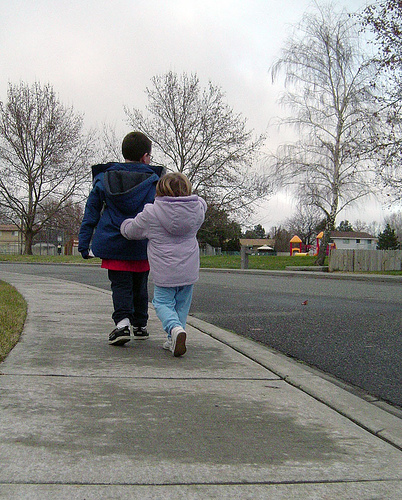}} & Original & $f(u, \textrm{Two kids \textcolor{ForestGreen}{walk home from school.}} \rightarrow \texttt{Neutral}$ \\
& Counterfactual   & $f(u, \textrm{Two kids are \textcolor{RedOrange}{walking down a sidewalk}.}) \rightarrow \texttt{Entail}$ \\
& Counterfactual  & $f(u, \textrm{Two kids are \textcolor{Purple}{going home from school}.}) \rightarrow \texttt{Neutral}$ \\

\noalign{\vskip 2mm}    
& \multirow{2}{*}{Explanation} & \textit{``Kids walking down a sidewalk are not necessarily going home from school.'' }\\ 
& & := \hspace{0.2em} $\neg(\textrm{\textcolor{RedOrange}{walking down a sidewalk}}$ \hspace{0.3em} $\Rightarrow$ \hspace{0.2em}  $\textrm{\textcolor{Purple}{going home from school}})$ \vspace{0.5em} \\


\end{tabular}
\caption{\small
Examples of counterfactual hypothesis rewrites from the explanation in SNLI-VE dataset \cite{do2020snli}. $f(u, x^{}) \rightarrow y$ shows the expected model prediction given the picture premise ($u$) and either original or counterfactual hypothesis. Note that \texttt{Neutral} can still result in \texttt{Neutral} (\autoref{sec:satisfiability}: Proposition 3). Colors are a visual guide for the relevant text spans which are used to construct the counterfactual.} 

\label{tab:examples}
\end{table*}
 
\subsection{Satisfiabilty of the Counterfactual}
\label{sec:satisfiability}
In this section, we derive what the counterfactual hypothesis and associated counterfactual label would be for each original label assuming logical formulas and discrete variables where exact substitution is possible. \autoref{sec:automatic} describes our suggested approach to handle Natural Language where discrete substitution no longer holds.

\begin{axiom}
Substitution for formulas \cite{fitting2012first} For any variables A and B and any formula $x_{[A]}$ containing $A$, if $x_{[B \backslash A]}$ is obtained by replacing any number of free occurrences of A in $x$ with $B$, then for $A \Leftrightarrow B$, $x_{[A]} \Leftrightarrow x_{[B \backslash A]}$.
\end{axiom}

\begin{assumption}
The hypothesis $x$ and premise $u$ contain $n$ free variables, of which the variables $A$ and $B$ are members in $u$ and $x$. We denote the membership of $A \subseteq x$ as $x_{[A]}$.

\end{assumption}
\begin{assumption}
 Given $R(A,B)$, the counterfactual hypothesis can be constructed by applying Axiom 1 replacing $A$ with $B$, denoted $x_{[B \backslash A]} = x^{cf}$.\footnote{Equivalence formulas may be substituted for one another without changing that formula’s truth value \cite{fitting2012first}.}
\end{assumption}

\begin{assumption}
The predicate relation expressed in $R(A,B)$ is a sufficient condition, to explain the model's predicted label.
\end{assumption}

We formally derive the associated counterfactual label $y^{\textrm{cf}}$ (expected satisfiability result) of the new counterfactual hypothesis $x^{\textrm{cf}}$. \autoref{fig:fig1} and \autoref{tab:examples} show examples for this process. 
The proofs follow the following high-level structure: 

\begin{enumerate}[wide, labelindent=0pt,itemsep=1pt,topsep=0pt]
    \item Substitution of variables $A$ and $B$ (Axiom 1 and Assumption 1) to construct $x^{\textrm{cf}}$ (Assumption 2).
    \item  Examine the predicate relationship $R(A,B)$ and derive the logical relationship between the $x^{\textrm{cf}}$ and the premise $u$.
\end{enumerate}

\begin{proposition}  
If the original label is \texttt{E}, then the associated counterfactual label is \texttt{E}. 
\end{proposition} 

\begin{proof} 
By Assumption 1, the logical proposition represented by \texttt{E} is $u \Rightarrow x_{[A]}$. Since $A \Leftrightarrow B$, then $u \Rightarrow x_{[B\backslash A]}$, by Axiom 1, $u \Rightarrow x^{\textrm{cf}}$. Therefore we have the resulting propositional formula and associated counterfactual label $(u \Rightarrow x) \Leftrightarrow (u \Rightarrow x^{\textrm{cf}})$, i.e. ($y=\texttt{E}) \Leftrightarrow (y^{\textrm{cf}}=\texttt{E})$.
\end{proof}

\begin{proposition}
If the original label is \texttt{C}, then the associated counterfactual label is \texttt{E}.
\end{proposition}

\begin{proof}
By Assumption 1, the logical proposition represented by \texttt{C} is $(u \Rightarrow \neg x_{[A]})$. Since $\neg(A \Leftrightarrow B)$ is equivalent to $(A \Leftrightarrow \neg B)$, then by Axiom 1, $u \Rightarrow \neg(x_{[\neg B \backslash A]})$. However it is not possible to test for the ``negation of'' variables as negation is not seen in training data for NLI. Under Assumption 3,\footnote{The violation of Assumption 3, can result in partially `unfaithful' explanations that do not provide enough information to explain model prediction.} if the explanation $\neg(A \Leftrightarrow B)$ sufficiently explains the label, then having $x_{[B \backslash A]}$ negates (or `flips') the label to Entailment $u \Rightarrow \neg\neg(x_{[B \backslash A]})$. Therefore we have the result $(u \Rightarrow \neg x) \Leftrightarrow (u \Rightarrow \neg \neg (x^{\textrm{cf}}))$, i.e., ($y=\texttt{C}) \Leftrightarrow (y^{\textrm{cf}}=\texttt{E})$.
\end{proof}


\begin{proposition}
If the original label is \texttt{N}, then there are \textbf{two} associated counterfactual labels, \texttt{E}, and \texttt{N}.
\end{proposition}

Two conditions arise because $\neg(A \Rightarrow B) \Leftrightarrow A$ AND $\neg B$. In the interest of space, we present the proof for Proposition 3 in the Appendix. \autoref{tab:cf-hypothesis} summarises the original propositional formula in the hypothesis $x$ and premise $u$, the predicate relations $R(A,B)$ and the associated
$y^{\textrm{cf}}$.


\subsection{Proposed Metric: FTC}

\label{sec:metric}
We introduce the metric Faithfulness-Through-Counterfactuals (FTC), to capture the difference in model predicted probabilities $p(\hat{y}^{\textrm{cf}})$ from the associated counterfactual label $y^{\textrm{cf}}$. 
\begin{equation}
\text{FTC} = 1-d(p(\hat{y}^{\textrm{cf}}), p(y^{\textrm{cf}}))
\end{equation}

\paragraph{Choice of Distance Function $(d)$} We consider three metrics for $d$, $\mathbbm{1}[\mathrm{argmax}(p(\hat{y}^{\textrm{cf}}))!=y^{\textrm{cf}}]$ denoted FTC-$\delta$, KL Divergence (FTC-$\mathcal{K}$) and Wasserstein distance (FTC-$\mathcal{W}$) with symmetrical distance of $1$ between \texttt{E} and \texttt{C}, and $0\geq \alpha \geq 1$ between \texttt{N} and the two other labels.

\subsection{Generating Counterfactual Hypotheses}
\label{sec:automatic}

In the previous section, we had assumed that the logical variables $A$ and $B$ are substitutable in $x$ directly. Indeed generating a counterfactual hypothesis would be trivial if $A$ and $B$ could be directly extracted from the explanation, and directly substituted in the hypothesis.\footnote{Consider the hypothesis: ``the boy is outside'' and the explanation: ``A tire swing is usually installed outside''. Naive substitution would result in ``the boy is a tire swing''.} However, open domain semantic parsing is an unsolved problem \cite{lee2021toward} of which to our knowledge, there is no off-the-shelf solution which does not require fine-tuning on a train set.\footnote{We experimented with dependency parses from spaCy and Stanford NLP which gave many irrelevant extractions.} Hence, we propose to leverage on advances in few-shot priming \cite{brown2020language} which only requires several handwritten examples and no further fine-tuning. This is compared against the baseline of parsing via `extract and transform' using regex (experiment described in \autoref{sec:validation_span}).

\begin{figure}[!t]
    \centering
    \includegraphics[width=0.485\textwidth]{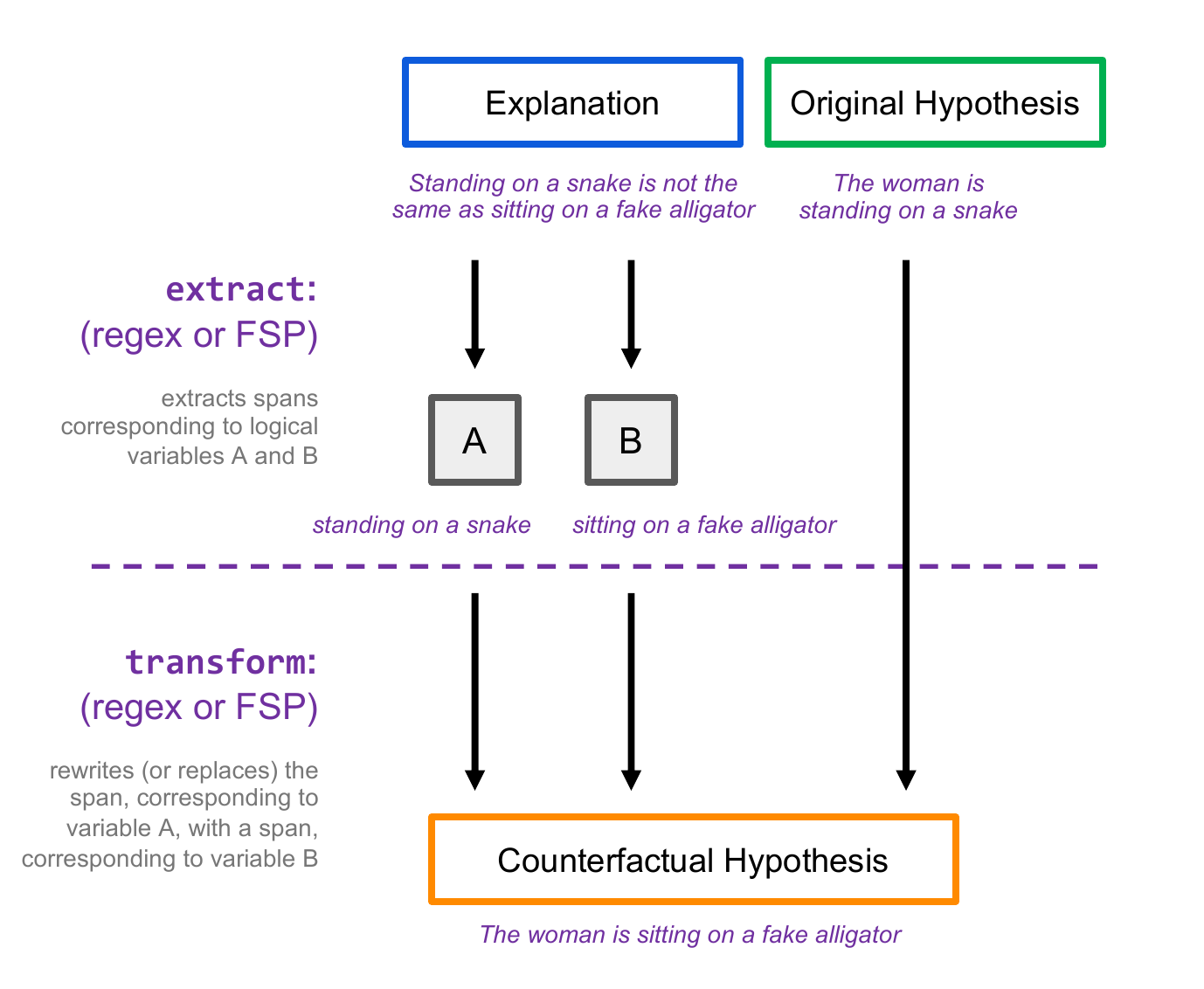}
    \caption{Two step counterfactual rewriting of the hypothesis, according to the explanation. We implement extract and transform steps using regular expressions (regex) and few-shot priming (FSP) models.}
    \label{fig:fig2}
\end{figure}

\paragraph{Extract and Transform with Regex}

We adopt and extend templates identified by \citet{camburu2019make} for explanations, who noted that these templates are a ``natural consequence of the task and dataset''. Regex methods perform a rule-based span extraction and replacement (we allow for stemmed word replacement). Extraction rules are shown in Appendix (\Cref{tab:regexe,tab:regexc,tab:regexn}).

\paragraph{Extract and Transform with Few-shot Priming}


We compare the brittle regex pipeline with the modern paradigm of in-context few-shot priming. This is an attractive option in our setting where we do not have prior training data for generating counterfactual hypotheses, and the task appears to be related to manipulation of text strings. 


We thus consider a two step process of 1) extracting the logical spans, i.e., $A$ and $B$ in \autoref{tab:cf-hypothesis} from the explanation, and 2) modifying the hypothesis given these extracted spans.\footnote{This two step-process is necessary as preliminary experiments show that even few-shot priming models with 175B parameters (academic access GPT3) are not able to construct counterfactual hypothesis in a single step.} 
Given a sequence of priming examples of how to extract spans from the explanation in the prefix, the model should perform the extraction given a test explanation.



\paragraph{Reducing Natural Language Artifacts with $x^{\textrm{cf}}$}
Previous attempts to test the explanation directly as input to a trained model, are subject to confounds of ``label leakage'' because of the close association between label and syntatic form of the explanation \cite{pruthi2020evaluating}. Crucially, our proposed method sidesteps this confound by applying \textit{logical satisfiability} checks via predicate relations $R(A,B)$. In theory, the construction of $x^{\mathrm{cf}}$ should preserve the syntatic structure of the original hypothesis, while only changing the semantics of $A$ and $B$. 

\section{Experiments and Results}
We validate our automatic method through three sets of experiments. 
\begin{itemize}
    \item[i)] Evaluating the quality of generated counterfactual hypotheses from a few-shot generator $\mathcal{H}^{\textrm{model}}$ (\autoref{sec:validation_span}).
    \item[ii)] Evaluating the proposed metric (FTC) on gold-model agreement of human generated $x^{\textrm{cf}}$, and comparing this to existing faithfulness metrics in the literature (\autoref{sec:validation_entailment}).
    \item[iii)]  Studying the sensitivity of our proposed approach compared to other metrics given pathological explanations (\autoref{sec:sensitivity_analysis}).
\end{itemize}

\subsection{Experimental Setup}

\paragraph{Datasets}
We consider logical entailment datasets, e-SNLI \cite{camburu2018snli} and e-SNLI-VE \cite{do2020snli} which are the only explainable logical entailment datasets available at point of writing \cite{wiegreffe2021teach}. e-SNLI consists of crowdsourced explanations for SNLI. e-SNLI-VE replaces the textual premise $u$ of SNLI with Flickr30k images \cite{young2014image}. 
To avoid trivial word overlap between $u$ and $x^{\textrm{cf}}$, we adopt the image representation for the premise $u$ in our experiments for \autoref{sec:validation_entailment} and \autoref{sec:sensitivity_analysis}. Note that \autoref{sec:validation_span} only requires $x$ and $z$.\footnote{All experiment code will be available at anonymous.io.}  


\begin{table*}[!t]
\centering
\resizebox{2.05\columnwidth}{!}{%

\small
\begin{tabular}{ll@{\hskip 0.3in}cccc@{\hskip 0.3in}cccc}
\toprule
  &         &        \multicolumn{4}{c}{Human Explanations $z^* \mapsto x^{\textrm{cf}}$ } & \multicolumn{4}{c}{Generated Explanations $\hat{z} \mapsto x^{\textrm{cf}}$ } \\
Extract & Transform & \texttt{C} & \texttt{E} & $\texttt{N}_{[A]}$ & $\texttt{N}_{[B]}$ &  \texttt{C} & \texttt{E} & $\texttt{N}_{[A]}$ & $\texttt{N}_{[B]}$ \\
\midrule
regex & regex &  0.690 & 0.638 & 0.010 & 0.012 & 0.744 & 0.701 & 0.473 & 0.418 \\ 
regex & Neo2.7B & 0.690 & 0.709 & 0.710 & \textbf{ 0.752} & 0.840 & 0.860 & \textbf{0.805} & \textbf{0.714} \\
Neo2.7B & regex & 0.801 & 0.691 & 0.509 & 0.565 & \textbf{0.893} & 0.781 & 0.669 & 0.584 \\
Neo2.7B & Neo2.7B & \textbf{0.822} & \textbf{0.782} & \textbf{0.743} & \textbf{0.754} & 0.870 & \textbf{0.881} & \textbf{0.807} & 0.709 \\
\bottomrule
\end{tabular}}

\caption{\small METEOR scores for hypothesis generated by regex, GPT-Neo, or their combination, vs human generated hypothesis. Bold is applied column-wise. We show the breakdown of revised hypothesis by class label (Contradiction \texttt{C}, Entailment \texttt{E} and Neutral \texttt{N}), and also whether the explanations are human or model generated. Note that there are two counterfactual hypothesis generated for \texttt{N}, $\texttt{N}_{[A]}$ and $\texttt{N}_{[B]}$ which correspond to $x^{\textrm{cf}}_{[A]}$ and $x^{\textrm{cf}}_{[B]}$ as described in \autoref{sec:satisfiability}: Proposition 3. }
\label{tab:validate_hyp}

\end{table*}

\paragraph{Models}
For the task model $f$, we adopt a state-of-art multimodal model, \texttt{CLIP} \cite{radford2021learning}, and fine-tune a 2-layer MLP to train a predictor $f(u, x) \rightarrow y$. For the explanation generator, $g$, we follow \citet{do2020snli} and fine-tune a modified GPT2 \cite{radford2019language}. 
Training details are in Appendix: \autoref{sec:experiment_details}.  
For the counterfactual hypothesis generator, $\mathcal{H}^{\textrm{model}}$ we adopt a pretrained GPT2-XL and GPT-Neo1.3B and 2.7B \cite{black2021gpt} without further fine-tuning, and apply only handwritten prompts. Prompt examples were randomly sampled from the training set and we used $20$ prompts for each label.\footnote{Further details are available in Appendix:\autoref{sec:experiment_details} and example of prompt templates(Appendix:\autoref{sec:prompt_templates})} 

\subsection{Quality of Counterfactual Hypotheses}
\label{sec:validation_span}

\begin{figure*}[!h]
    \centering
    \includegraphics[width=0.85\textwidth]{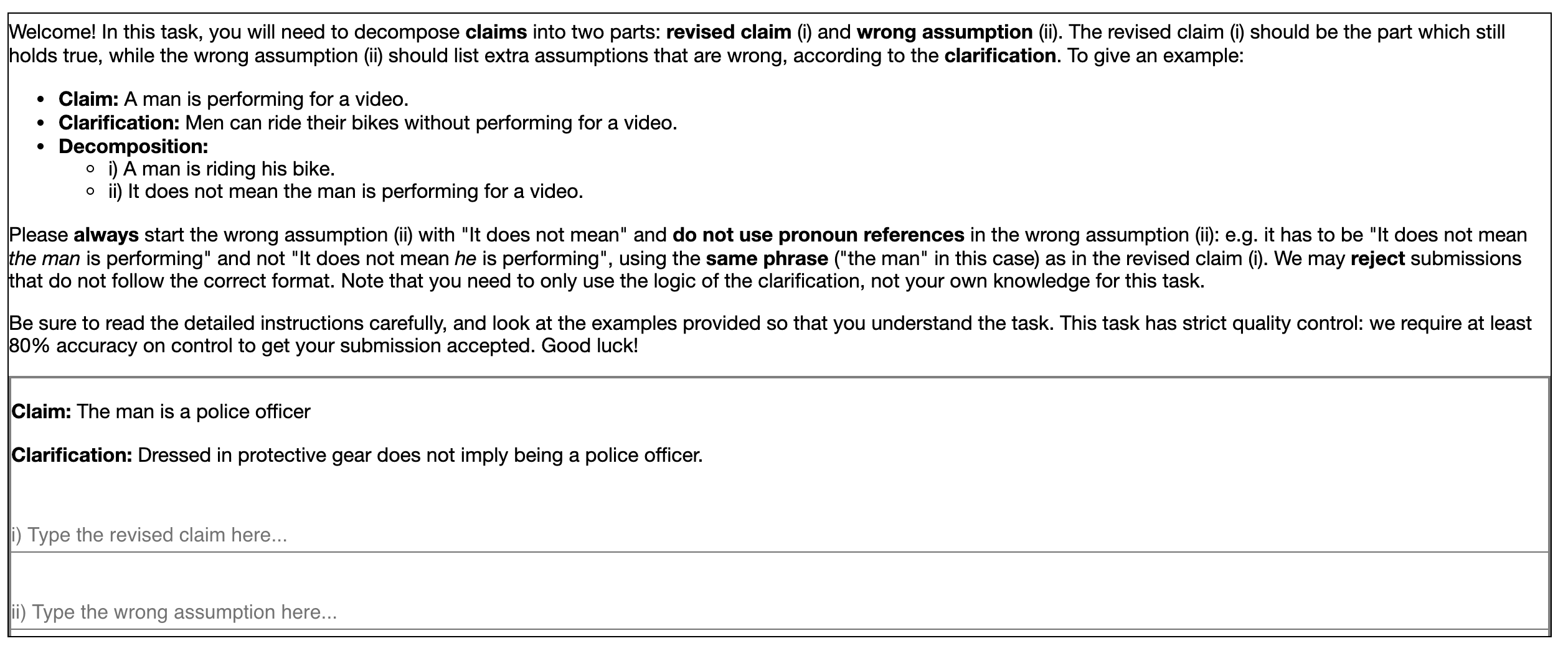}
    \caption{Interface shown to annotators in the counterfactual hypothesis rewriting task.}
    \label{fig:cf_rewrite_task1}
\end{figure*}

As the feasibility of our automatic approach depends on the quality of counterfactual generation, we evaluate $\hat{x}^{\textrm{cf}} \leftarrow \mathcal{H}^{\textrm{model}}(x, z)$ against gold counterfactuals, $x^{\textrm{cf}^*} \leftarrow \mathcal{H^{\textrm{human}}}(x, z)$, where $\mathcal{H^{\textrm{human}}}$ refers to the human annotator, $\mathcal{H}^{\textrm{model}}$ refers to our automated hypothesis generator, $z$ refers to explanations and $x$ refers to the original hypothesis. We randomly sample 300 examples from the validation set (100 each for \texttt{E}, \texttt{C}, \texttt{N}) and ask annotators to write counterfactual hypothesis for human generated explanations, $z^*$ and model generated explanations $\hat{z}$. 
Annotators are asked to revise $x$ such that the logic in $z$ is expressed in the new counterfactual $x^{\textrm{cf}}$ (Appendix:\autoref{fig:cf_rewrite_task1}). We show annotators the \textit{same} set of examples that were used to prompt $\mathcal{H}^{\textrm{model}}$. We obtain three annotations per datapoint for multiple reference sentences.\footnote{``A young boy wearing a white shirt on a beach'' and ``A young boy on a beach wearing a white shirt'' are both valid.} 

\paragraph{Experiment Conditions}
As described in \autoref{sec:automatic}, we adopt a two-step process of `extract' and `transform' for generating $\hat{x}^{\textrm{cf}} \leftarrow \mathcal{H}^{\textrm{model}}(x,z)$. We compare different combinations of either extracting and transforming with regex or $\mathcal{H}^{\textrm{model}}$. Our main results for the largest and best performing model (GPT-Neo2.7B) is shown in \autoref{tab:validate_hyp}, and additional results for GPT-Neo1.3B and GPT2-XL in Appendix: \autoref{tab:validate_hyp_appendix}. 

\paragraph{Metric}
The evaluation metric used is METEOR \cite{banerjee2005meteor}, that was found to correlate with human judgement \cite{kayser2021vil}. METEOR computes harmonic mean of unigram precision and recall and accounts for stemming. As the validity of any text generation metric is debatable \cite{deng-etal-2021-compression}, we further quantify the downstream effects of the automated process through \autoref{sec:validation_entailment}.

\paragraph{Results} (Presented in \autoref{tab:validate_hyp}).

\begin{enumerate}[wide, labelindent=0pt,itemsep=1pt,topsep=0pt]
    \item The combination of using models for both extract and transform steps (last row) performs best in all cases with human explanations $z^*$, and close to best with model explanations, $\hat{z}$. 
    \item The performance of most methods are better on $\hat{z}$ than $z^{*}$ which might be explained by the more `standardised' text format in $\hat{z}$. \citet{brahman2020learning} reported that generator models tend to follow a similar format, supporting this interpretation. The row regex-regex can be seen as a direct comparison of how `standardised' $\hat{z}$ is compared to $z^{*}$ as it indicates the performance for a brittle rule-based approach.  

    \item For ease of rewriting each class (column-wise), $\texttt{C} > \texttt{E} > \texttt{N}$ for $z^*$ in most cases, which highlight the relative complexity (\texttt{N} tends to be expressed in a less `straightforward' manner) of extracting and transforming hypothesis with different types of explanations.  
    
\end{enumerate}



\subsection{Metric Validation for Predictions on Counterfactual Hypotheses} \label{sec:validation_entailment}


 
 In this experiment, we compare faithfulness scores with human predictions on counterfactual hypotheses, assuming that humans are consistent with the logically derived answer (we describe the case when they are not in the next section). 
 
 As described in \autoref{sec:introduction}, faithfulness metrics should focus on utility of explanations to the user \cite{jacovi2020towards}. One practical utility is that humans should have a better handle of the model's predictions on slightly different input.\footnote{Simulatability studies \cite{doshi2017towards} using raw explanations were considered, however this is subject to confounds in Label leakage\cite{pruthi2020evaluating,hase2020leakage} due to a nearly one-to-one correspondence with the linguistic form of explanations and the label.} 
  The domain of new input that we test, is the counterfactual hypotheses $x^{\textrm{cf}} \leftarrow \mathcal{H}^{\textrm{human}}(z^*, x)$ that has been generated by humans using human explanations in \autoref{sec:validation_span}. Faithfulness metrics should therefore be good at discriminating whether the gold predictions agree with the model predictions. 
  

  
 Since faithfulness metrics are real-valued, and agreement is discrete binary, we are interested in how well they can discriminate between the two groups. We use the Wilcoxon rank-sum test which is a nonparametric test for the null hypothesis that two groups (agreement vs disagreement) are equal. Recall that our FTC metric proposes to use $\mathcal{H}^{\textrm{model}}$ to generate a counterfactual hypothesis from the explanations. If $\mathcal{H}^{\textrm{model}}$ had produced exactly the same $x^{\textrm{cf}}$ as the human $\mathcal{H}^{\textrm{human}}$, then FTC should be very effective at separating the two outcome groups, and the test statistic $\rho \in [0,1]$ would be very close to $1$. However we know this automated process is imperfect (\autoref{sec:validation_span}), and the implications of a $0.7$ or $0.8$ METEOR score are unclear.
 
\begin{figure*}[!h]
    \centering
    \includegraphics[width=0.85\textwidth]{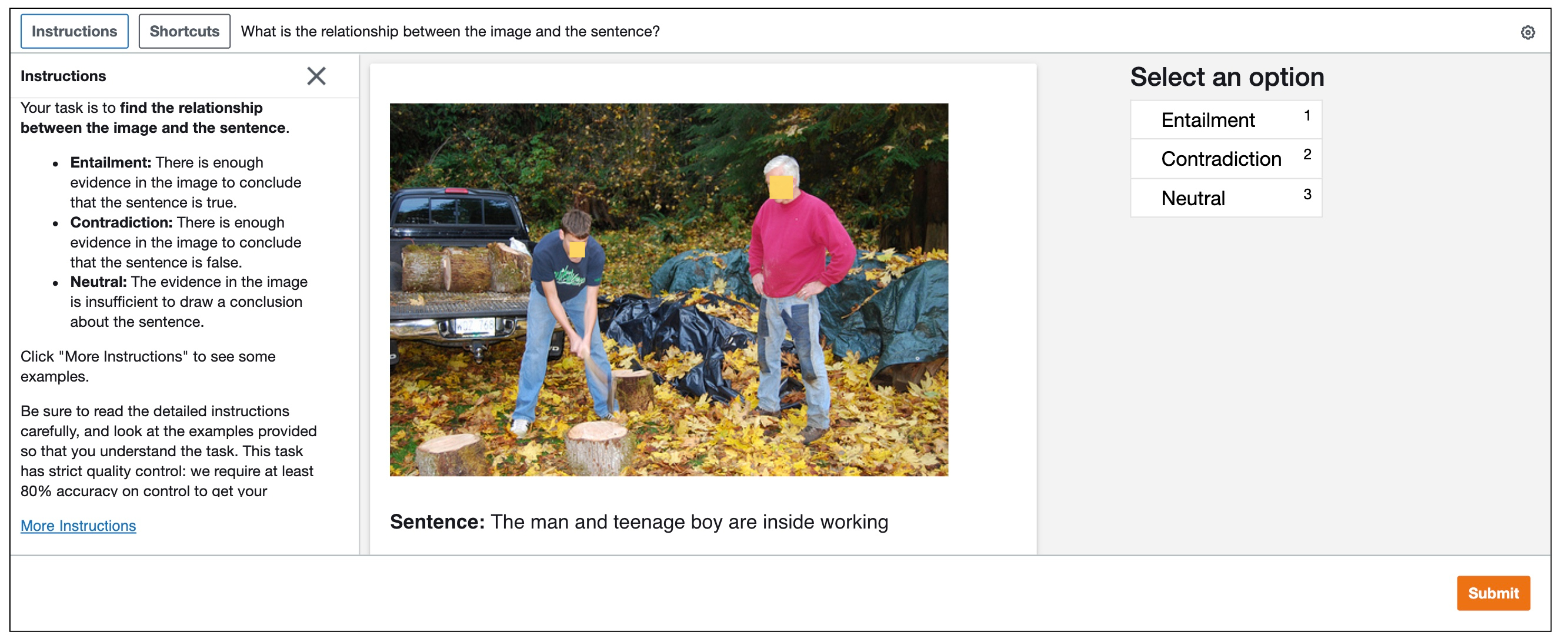}
    \caption{Interface shown to annotators in the NLI label collection task, mirroring the original SNLI protocol.}
    \label{fig:nli_task2}
\end{figure*}

We collect annotations for $100$ data instances grouped by each original label-class, and obtain $3$ annotations per instance.\footnote{The inter-annotator agreement, as measured by Fleiss' kappa, equals $0.51$. We aggregate the final label using the majority vote.}
Annotators are required to rate whether ($x^{\textrm{cf}}, u$) entails, contradicts, or is neutral (Appendix:\autoref{fig:nli_task2}). We compare with other faithfulness metrics namely Label Adjusted Simulation (LAS; \citet{hase2020leakage}) and Label Rationale Association (LRA; \citet{wiegreffe2020measuring}), reviewed in \autoref{sec:related_work} and Appendix: \autoref{tab:literature_review}. 

\begin{table}[!t]
\small
\resizebox{\columnwidth}{!}{
\begin{tabular}{l@{\hskip 0.25in}llll}
 & \texttt{C} & \texttt{E} & $\texttt{N}_{[A]}$ & $\texttt{N}_{[B]}$ \\
\toprule

LAS & 0.585 & 0.481 & 0.436 & 0.410 \\
LRA & 0.547 & 0.645* & 0.400 & 0.572 \\
FTC-$\delta$ & 0.627 & 0.634* & 0.581 & 0.655* \\
FTC-$\mathcal{K}$ & 0.743* & 0.747* & 0.631 & 0.644* \\
FTC-$\mathcal{W}$ & 0.692 & 0.748* & 0.605 & 0.644* \\

\end{tabular}}
\caption{\small $\rho$-statistic $\in[0,1]$ for Wilcoxon rank-sum test on different faithfulness metrics. A larger $\rho$-statistic indicates a larger effect size. LAS: Label Adjusted Simulation, LRA: Label Rationale Agreement, and FTC: Faithfulness-by-Counterfactual (ours), and FTC-$\mathcal{W}$ and FTC-$\mathcal{K}$ ($\alpha=0.7$) are the Wasserstein and Kl-divergence variants. * indicates significance (p-value$<0.05$). The test is conducted between the two groups; data points where the human agrees vs disagrees with the model's prediction on counterfactual hypothesis. }
\label{tab:ranksum}
\end{table}

\paragraph{Results} (Presented in \autoref{tab:ranksum})
\begin{enumerate}[wide, labelindent=0pt,itemsep=1pt,topsep=0pt]
    \item FTC variants have the highest $\rho \in [0, 1]$ statistic (higher is better and indicates a larger effect size), which indicates that it is the most discriminative metric for whether the human can simulate the model's prediction given new counterfactual inputs. This is expected as the simulation procedure is similar to how FTC is calculated. 
    
    \item As reported in \autoref{sec:validation_span} discrepancies in the automated rewriting process affect the scoring of the metric in ways which are not easily captured by sentence generation metrics (\autoref{tab:validate_hyp}). The results show that the best-automatic rewriting process of adopting GPT-Neo2.7B achieves $0.58$ to $0.75$ $\rho$-statistic across \texttt{C, E, N}, giving us an indication of the downstream impact of $0.74$ to $0.82$ METEOR score. 
    
    \item KL-Divergence (FTC-$\mathcal{K}$) and Wasserstein Distance (FTC-$\mathcal{W}$) perform slightly better than the naive Identity function (FTC-$\delta$). This indicates that a soft computation over probabilities can account for some of the errors in the $x^{\textrm{cf}}$ rewriting process. 
    \item  With the exception of LAS on $\texttt{C}$ and LRA on $\texttt{E}$, we find that other metrics have poorer $\rho$ across $\texttt{C, E, N}$. This indicates that the metric is relatively poorer at distinguishing between the two outcome groups.  
\end{enumerate}

\paragraph{Examining Inconsistent Explanations}

In cases where the human NLI label does \textit{not} correspond to the logically derived NLI class in \autoref{tab:cf-hypothesis}, our method suggests that the human provided explanations themselves are \textit{not} faithful to the human's `internal NLI model'.  Appendix:\autoref{sec:low_quality_explanations} shows examples of these cases, which we typically find to be due to explanations of low quality, supporting the central thesis of the paper. In the previous results, we filter out data points where majority of the annotators do not predict the logically derived NLI class. We report the non-filtered results in  \autoref{sec:low_quality_explanations}) which demonstrate negative results and a failure case when the `gold' human explanations are themselves problematic.\footnote{Quantifying the extent and range of annotation errors for explanations is outside of the scope of this work and we refer readers to \citet{valentino2021natural} who report that explanations in SNLI are valid logical relations 60\% of the time and other times they are either redundant or non-sensical due to annotation errors.
}

\subsection{Sensitivity Analysis}
\label{sec:sensitivity_analysis}
 
As a sanity check, good faithfulness metrics should be sensitive towards unfaithful explanations, i.e. they should perform worse on unfaithful explanations compared to faithful ones. We perform a sensitivity analysis on various faithfulness metrics by examining their raw 
scores on unfaithful explanations \textit{by construction}. These are constructed by leaving out all but one type of input to the explanation generator. The `complete' set of inputs are the entailment label ($y$), hypothesis ($x$), and premise ($u$). Note that the `complete' set of inputs to the explanation generator does not guarantee faithful explanations, but they are guaranteed to be less pathological than leaving out all but one type of input. 
We additionally consider BertScore \cite{zhang2019bertscore} and METEOR \cite{banerjee2005meteor} which are text similarity metrics evaluated against the human explanation, and use FTC-$\mathcal{K}$ variant which has an upperbound of $1$ and a high $\rho$-statistic in the previous experiment.\footnote{As described in \autoref{sec:metric}, FTC-$\mathcal{K}$ is $1- \textrm{the KL term}$, which has lower bound $0$ and no upper bound. Hence the upper bound on FTC-$\mathcal{K}$ is 1.}


\begin{table}[!t]
\small
\centering
\resizebox{\columnwidth}{!}{
\begin{tabular}{p{0.2cm}p{0.2cm}p{0.3cm}@{\hskip 0.2in}rrrrr}
$x$ & $u$ & $y$ &  BER & MET  &  LAS & LRA & FTC-$\mathcal{K}$ \\
\toprule
\checkmark  & \checkmark  & \checkmark &   0.888 &   0.245 &  0.047 &  0.788 &  0.126 \\
\checkmark  & {}  & {} & 0.885 &   0.233 & -0.035 &  0.561 & -0.200 \\
{}  & \checkmark  & {} & 0.869 &   0.120 & -0.063 &  0.298 & -0.055 \\
{}  & {} & \checkmark  &  0.865 &   0.097 &  0.068 &  0.630 & -0.245 \\


\end{tabular}
}
\caption{Raw scores for different metrics, by perturbing inputs to the explanation generator (sensitivity analysis). $x$, $u$, and $y$ refers to hypothesis, premise, and label respectively. BER: BertScore, MET: METEOR, LAS: Label Adjusted Simulation, LRA: Label Rationale Association, FTC-$\mathcal{K}$ (ours). Note that the first two are measures of text similarity, while the other three are designed to measure the faithfulness.}
\label{tab:not_faithful}
\end{table}

\paragraph{Results} (presented in \autoref{tab:not_faithful})
\begin{enumerate}[wide, labelindent=0pt,itemsep=1pt,topsep=0pt]
    \item FTC-$\mathcal{K}$ performs consistently better for the  non-pathological (first row $0.126$) vs pathological explanations ($-0.200$, $-0.055$, $-0.245$). The same `correct' trend is observed for LRA $0.788$ vs ($0.561$, $0.298$, $0.630$). 
    \item We find that an off-the-shelf BertScore \cite{zhang2019bertscore} has a suprisingly low range of values for the different conditions ($0.865$ to $0.888$). METEOR scores conditioned only on $x$ are also very close to the full range of inputs ($0.233$ vs $0.245$) indicating superficial word similarity of the explanations when conditioned on just $x$.
    
    \item LAS scores are in the `wrong' direction, namely that explanations generated with all of the relevant inputs perform worse than just having the label. 

\end{enumerate}

\section{Related Work}
\label{sec:related_work}


We outline existing methods which evaluate free-text generated Natural Language explanations, their assumptions of faithfulness, and describe how they operationalise these assumptions. We focus on LAS and LRA, which are used in our experiments, and provided more discussion of related work in Appendix:\autoref{sec:related_work_appendix}. 


\subsection{Leakage Adjusted Simulation}

This method assumes explanations are faithful if they allow a model to be more simulatable. A model is simulatable to the extent that an observer, or simulator, can predict its outputs 
\cite{doshi2017towards, hase2020leakage,kumar2020nile}. From this perspective, one might use a simulator’s (either a human or model) accuracy with explanations as input, to measure explanation quality. However, as \citet{hase2020leakage} argues, the simulator’s success does not reflect explanation quality when the explanation leaks the label to the simulator. They thus propose Leakage-adjusted Simulatability (LAS) which performs a macro-average of leakable and non-leakable explanations. However, a high occurrence of label leakage may overwrite the effect of macro-averaging \cite{pruthi2020evaluating}. 


\subsection{Label Rationale Association}

According to \citet{wiegreffe2020measuring}, ``at a minimum, rationales must be implicitly or explicitly tied to the model’s prediction.''. Their method tests whether label and explanations are similarly robust to noise in the input. Although designed to be highly generalisable to generative framework of explanations, this assumption may be overly general for more rigorous notions of faithfulness. Consider the scenario where merely changing the label to the generator results in ``sufficiently different'' explanations being generated \cite{kumar2020nile}, whether or not the original explanation was actually faithful, LRA will assign this a high score.

\subsection{Counterfactuals as Explanations}

Our approach differs from the literature on Counterfactuals as explanations \cite{mothilal2020explaining, verma2020counterfactual} as we do not generate counterfactual explanations, but generate counterfactual hypotheses based on the explantions. \citet{camburu2019make} work has a similar flavor, where they ``reverse” a hypothesis. However, they focus on show the pathologies of a generator by searching for (adversarial) input hypothesis that cause the model to generate logically inconsistent explanations. \citet{ge2021counterfactual} also constructs `counterfactual inputs', but search for existing features in the original input, and consequently is only applicable to extractive explanations. 

 
\subsection{Natural Logic vs Free Logic}

Previous work on ``Natural Logic'' \cite{maccartney2007natural} relies on natural language features to guide inferences. 
For instance, changing specific terms to more general ones preserves entailment. This sidesteps the difficulties of translating sentences into First-order-Logic. Natural logic systems \cite{angeli2014naturalli} have been used in explainable fact verification \cite{krishna2021proofver} which constructs ``explanations’’ by presenting logical steps for inference. However these approaches still require a knowledge base to train or mine truth values, e.g, ``in Paris” $\subseteq$ ``in France”. In contrast, our method does not require additional training and is a procedurally lightweight method relying on off-the-shelf pretrained models.

\section{Conclusion}
Measuring faithfulness of free-text explanations with respect to a task model is a challenging problem due to confounds introduced by testing explanations directly. In this work, we propose an approach to evaluating explanations for NLI tasks which uses the predicate logic expressed in explanations to construct counterfactual hypotheses, and tests the \textit{satisfiability} of the resulting hypothesis. Our experiments on validating counterfactual hypothesis generation and the fully automated approach distinguishing when humans agree and disagree with the model's prediction on counterfactual hypotheses show that a proposed automatic pipeline is a viable approximation to the theoretical method. Further, we show that our metric is sensitive to pathologically unfaithful explanations.     


\section{Impact Statement}
\subsection{Potential Limitations}
\label{sec:limitations}
We document and discuss some potential limitations of our proposed approach.
\begin{enumerate}[wide, labelindent=0pt,itemsep=1pt,topsep=0pt]
    \item The suggested implementation (\autoref{sec:automatic}) relies on pre-trained models which are open-source,\footnote{\url{https://huggingface.co/EleutherAI/gpt-neo-2.7B}} but requires GPU memory of $>$10Gb memory to host the 2.7B parameter model and $<$10Gb to host the 1.3B parameter model.
    
    \item While it is reasonable to expect that applying the same approach using larger GPT models might generate better counterfactual hypotheses, we have opted to use models that can still be launched on a single GPU machine, and does not assume access to industry level 175B models. 
    
    \item Our metric is not directly differentiable as it involves discrete operations to generate the counterfactual hypotheses.
    
    \item Since the method of generating counterfactual hypotheses is tied to the set of explanations, the generated counterfactual hypotheses are not meant to be an extensive set of counterfactuals that might be used to test a model, which has been explored in other work such as \citet{ross-etal-2021-explaining}. This is not strictly a limitation, but a note that the method focus on evaluating generated explanations, rather than generating counterfactuals extensively.
    
    \item While most classification problems can be reformulated into ternary logic of $\texttt{E, C, N}$, it is more challenging to extend the notion of logical \textit{satisfiability} for tasks that require open-ended text generation, and so the current work is limited to NLI task explanations.
\end{enumerate}

\subsection{Potential Risks}

The proposed work investigates an evaluation methodology for faithfulness of explanations. This work focuses on explanations for the NLI task, which has no influence on the actual classification algorithm of the NLI model. Extensions of the method which generate counterfactual hypothesis using other pretrained models may be subject to hallucinations in those models (which we did not currently encounter with the present models). 


\section*{Acknowledgments}
We thank Benjamin Van Durme for  computational resources for experiments that were ran outside Meta AI. Alexandra Delucia, Marc Marone, Patrick Xia, Matthew Francis-Landau, Lin Chu-Cheng for comments and discussion. 

\bibliography{anthology,custom}
\bibliographystyle{acl_natbib}

\clearpage

\section{Appendix}
\subsection{Notation Table}
\begin{table}[!h]
\small

\begin{tabular}{ll}
\small
$f$ & task model (CLIP) \\
$g$ & explanation generator (GPT2) \\
$\mathcal{H}$ & counterfactual hypothesis generator \\
$\mathcal{H}^{\textrm{model}}$ & counterfactual hypothesis generator (GPT-Neo) \\
$\mathcal{H}^{\textrm{human}}$ & counterfactual hypothesis generator (human) \\
$z^{*}$ & explanations (human) \\
$\hat{z}$ & explanations (generator $g$) \\
$u$ & premise \\
$x$ & hypothesis (original) \\
$x^{\textrm{cf}}$ & hypothesis (counterfactual) \\

$y^{*}$ & task label (human) \\
$\hat{y}$ & task label (model $f$) \\
$y^{*cf}$ & task label on counterfactual (expected) \\
$\hat{y}^{cf}$ & task label on counterfactual (model $f$)  

\end{tabular}
\caption{\small
Summary of notation used in the paper
}

\label{tab:notation}
\end{table}

\subsection{Satisfiability of the Counterfactual: Proposition 3}
\textit{If the original label is \texttt{N}, then there are \textbf{two} associated counterfactual labels, \texttt{E}, and \texttt{N}.}

For \texttt{N}, as $u \xRightarrow{(\textbf{?})} x$  is truth-valueless under Open Universe Semantics \cite{russell2015unifying}), the formula is either truth-valueless of every variables in 
$x$ and $u$, or truth-valueless of some and false of others \cite{sep-logic-free}. We start by considering the predicate relation $R(A,B) = \neg (A \Rightarrow B)$ which we rewrite as $A \wedge \neg B$, so that the two variables are separately testable (due to $\wedge$ relation).

For $A$, applying Assumption 1, we verify the satisfiability of $u \Rightarrow x^{\textrm{cf}}_{[A]}$ where the counterfactual hypothesis is $x^{\textrm{cf}}_{[A]} = x_{[A\backslash B]}$. We test the formula which excludes $B$ but has $A$. The associated counterfactual result of $y^{\textrm{cf}}_{[A]}=\texttt{E}$.


Next we consider $\neg B$, with the counterfactual hypothesis $x^{\textrm{cf}}_{[\neg B]} = x_{[\neg B \backslash A]}$. Since the original propositional formula was truth-valueless, the same conditions apply for $u \xRightarrow{\textbf{(?)}} x_{[\neg B]}^{\textrm{cf}}$ which is still truth-valueless. The negation of a truth-valueless formula is still truth-valueless \cite{sep-logic-free}) hence  $u \xRightarrow{\textbf{(?)}} x_{[B]}^{\textrm{cf}}$ has the counterfactual label $y^{\textrm{cf}} = \texttt{N}$.

\subsection{Experiment Details}
\label{sec:experiment_details}

\paragraph{Dataset} 
The dataset contains a reduced sample of the
original 570k sentence pairs from SNLI where \citet{do2020snli} apply various filtering methods to
remove noise that occurred from combining e-SNLI and
SNLI-VE. The training/val/test splits are 401718/14340/14741. Additional details about e-SNLI-VE including distribution are available at \url{https://openaccess.thecvf.com/content/ICCV2021/supplemental/Kayser_E-ViL_A_Dataset_ICCV_2021_supplemental.pdf}. Our use of this dataset is compatible with original access conditions and in research contexts.

\paragraph{Package Details}
For all models, we used Huggingface's \texttt{Transformers, v 4.12.2}. For Meteor Score calculation, we used \texttt{NLTK v 3.6.1} which lower cases tokens. For BertScore calculation, we used \texttt{roberta-large} and the metric implementation from Huggingface's \texttt{datasets v 1.15.1}. For training and validation loop, we used \texttt{PyTorch Lightning, v 1.5.2.}

\paragraph{Training Details}
Models are trained with Adam Optimizer \cite{kingma2014adam}  with learning rate $10^{-5}$ and batch size 64. Models were trained using with Early stopping was used with \texttt{patience=5} and \texttt{threshold=0.0001}. We did not perform any hyperparameter search for this work, and used default values for training the models from the packages. 

\paragraph{Computational Budget}
The largest model used was GPT-Neo 2.7B (2.7 Billion Parameters). The GPU hardware for the experiments used was \texttt{NVIDIA's Quadro RTX 6000}. The multimodal CLIP model takes about a day to fine-tune. To generate $14000$ new counterfactual hypothesis using pretrained GPT-Neo 2.7B takes about 1-2 hours.

\paragraph{Prompt Selection Details}
We report results from a single run of random prompt selection, as hand-labelling multiple random prompt sets is manually intensive. A preliminary experiment comparing randomly sampled prompt sets, versus obtaining prompts using datapoints that are closest to the cluster centers from k-means clustering (where k=20)  of the sentence unigram and bigram vectors, yielded nearly identical scores for Table 3. With a huge source dataset to sample from, we suspect that both types of prompt set selection might be close to random.

\subsection{Related Work}
\label{sec:related_work_appendix}
\paragraph{Leakage Adjusted Simulatability}

For task inputs $X=\{x_i\}$, model outputs $\hat{Y}=\{\hat{y}_i\}$, and model explanations = $\hat{E}=\{\hat{e}_i\}$, Leakage Adjusted Simulatability (LAS) metric is computed as:

\begin{align*}
    \mathrm{LAS}_0 &= \frac{1}{n_0} \sum_{i:k_i=0}(\mathbbm{1}[\hat{y}_i \mid x_i, \hat{e}_i] - \mathbbm{1}[\hat{y}_i|x_i])\\
    \mathrm{LAS}_1 &= \frac{1}{n_0} \sum_{i:k_i=1}(\mathbbm{1}[\hat{y}_i \mid x_i, \hat{e}_i] - \mathbbm{1}[\hat{y}_i|x_i])\\
    \mathrm{LAS} &= \frac{1}{2}(\mathrm{LAS}_0 + \mathrm{LAS}_1) 
\end{align*}

where $k_i=\mathbbm{1}[\hat{y}_i \mid \hat{e}_i]$ is a leakage indicator, and $n_0$ and $n_1$ are the number of examples in non-leaking and leaking groups respectively \cite{hase2020leakage}.

Following \cite{hase2020leakage}, we randomly drop out either $\hat{e}_i$ or $x_i$ during training and can obtain the relevant conditional $\mathbbm{1}[\hat{y}_i \mid x_i, \hat{e}_i]$, $\mathbbm{1}[\hat{y}_i \mid x_i]$, or $\mathbbm{1}[\hat{y}_i \mid \hat{e}_i]$ during inference time to compute $\mathrm{LAS}$. We adopt the same set of final hyperparameters for dropout, $0.2$ explanations only, $0.4$ hypothesis and premise only, and $0.4$ hypothesis, premise and explanations. 

\paragraph{Label Rationale Association}
We adopt the ``Robustness Equivalence measure'' which tests whether both label and explanations are similarly or disimilarly robust to noise in the input. Following \cite{wiegreffe2020measuring}, we measure changes in label prediction as the number of predicted dev set labels that flip, i.e., change from their original prediction to something else, and measure changes in rationale quality using simulatability. Their approach aims to measure association between input perturbation and output explanation. They first add zero-mean Gaussian noise $\mathcal{N} (0, \sigma^2 )$ to each input embedding at inference time, and measure the changes in label prediction by counting the number of predicted labels in the test set which flip. They then use the corresponding generated explanations, and get the task prediction change of a separate model that has been pretrained with explanations.  The LRA is computed as:

\begin{align}
    F_i &= \mathbbm{1}[(\hat{y}_i|x_i)=(\hat{y}_i|x'_i)]\\
    Z_i &= \mathbbm{1}[\hat{y}_i \mid x_i, \hat{e}_i] - \mathbbm{1}[\hat{y}_i|x_i]\\
    \mathrm{LRA} &= \frac{1}{n} \sum_i^n \mathbbm{1}[F_i= Z_i]
\end{align}


\paragraph{Teacher-Student Paradigm}
\citet{pruthi2020evaluating} introduce a student-teacher paradigm, where they target a notion of faithfulness being tied to the usefulness of explanations. They measure the extent to which explanations allow student models to simulate the teacher model on unseen examples for which explanations are unavailable. Student models incorporate explanations in training (but not prediction) procedures. The proposed framework for evaluating explanation quality suggests that explanations are effective if they help students learn about the teacher.

\paragraph{Erasure}
The vast majority of prior work that proposes models and evaluations of explanatory methods for faithfulness focus on removing tokens from the input text \cite{deyoung2019eraser}. This method directly intervenes in the input hypothesis by finding `important' spans and is not applicable to free-text explanations. Our method has a similar spirit of doing direct interventions. However instead of removing tokens, we generate a counterfactual hypothesis which is grammatically close to the original hypothesis, but with propositions from the explanation. 

\onecolumn

\begin{table*}[!h]
\small
\small

    \begin{tabular}{p{13em}p{4em}p{16em}p{11em}}
 & Tests task model & Reliance on Third Party & Resources Required  \\
 \toprule
HS \cite{doshi2017towards} & No & Human verification & Human annotators  \\
LAS \cite{hase2020leakage} & No & Model verification & Training with explanations  \\
TS \cite{pruthi2020evaluating} & No & Model verification & Training with explanations  \\
LRA \cite{wiegreffe2020measuring} & No & Model verification &  Training with explanations \\
Faithfulness-through-Counterfactuals & Yes & Counterfactual hypothesis generator & Manual prompts

\end{tabular}



\caption{Comparison of existing faithfulness metrics for free-text explanations. HS: Human Simulation, LAS: Leakage Adjusted Simulation, TS: Teacher-Student, LRA: Label Rationale Association. Compared to the other models, our proposed method 'Faithfulness-through-Counterfactuals' directly tests the task model without relying on third party Human or model verification. The reliance on third party is via counterfactual hypothesis generator, which we validate the efficacy in \autoref{sec:validation_span}. Additionally, our method relies a one-off manual prompts generation which is cheaper than gathering explanation training data or relying on human annotators.}

\label{tab:literature_review}
\end{table*}

\subsection{Annotation Details}\label{sec:annotation_deets}
We recruited annotators from the pool of Amazon Mechanical Turk workers who are located in the United States. In the counterfactual hypotheses rewriting task (\autoref{sec:validation_span}), we paid between \$0.15 (for \texttt{E} and \texttt{C}) and \$0.20 (for \texttt{N}) per single rewrite, estimating that a sufficiently experienced worker can perform the task in under 45 seconds, thus yield a minimum of \$12 hourly rate. (We do not separately ask humans to extract logical spans.) In the NLI label collection task (\autoref{sec:validation_entailment}), we paid \$0.10 per single NLI label and estimated that each task should take a sufficiently experienced worker at most 30 seconds, thus yielding a minimum of \$12 hourly rate. 

\clearpage
\twocolumn


\subsection{Low Quality Explanations}
\label{sec:low_quality_explanations}
We find that inconsistency between human NLI labels and the logically derived labels can be largely attributed to low quality human explanations from the dataset. The logically derived label is obtained from \autoref{tab:cf-hypothesis}. Human NLI labels do not correspond to the derived counterfactual around $15$ to $45$\% of the time, with the following breakdown.

\begin{table}[!h]
\resizebox{\columnwidth}{!}{
 \small
\begin{tabular}{l@{\hskip 0.25in}cccc}
$\texttt{C} \mapsto \texttt{E}$ & $\texttt{E} \mapsto \texttt{E}$ & $\texttt{N}_{[A]} \mapsto \texttt{E}$ & $\texttt{N}_{[B]} \mapsto \texttt{N}$ \\
    \toprule
      46.6 & 13.3
 & 24.4 & 28.6 
\end{tabular}
}
\caption{\% of mismatches between human predicted and logically derived NLI label (RHS of $\mapsto$; \autoref{sec:problem_formulation}).}
\label{tab:agreement}
\end{table}

\autoref{sec:validation_entailment} presents results for filtering out data points where the majority label from the annotators is consistent with the logically derived NLI class. We additionally report non-filtered results in \autoref{tab:ranksum_uf} which reflect the failure case when the human (gold) generated explanation itself is not reliable.  

\begin{table}[!h]
\small
\resizebox{\columnwidth}{!}{
\begin{tabular}{l@{\hskip 0.25in}llll}
 & \texttt{C} & \texttt{E} & $\texttt{N}_{[A]}$ & $\texttt{N}_{[B]}$ \\
\toprule

LAS & 0.570 & 0.483 & 0.491 & 0.449 \\
LRA & 0.547 & 0.561 & 0.509 & 0.602* \\
FTC-$\delta$ & 0.438 & 0.530 & 0.509 & 0.659* \\
FTC-$\mathcal{K}$ & 0.448 & 0.579 & 0.437 & 0.645* \\
FTC-$\mathcal{W}$ & 0.465 & 0.602* & 0.512 & 0.645* \\

\end{tabular}}
\caption{\small $\rho$-statistic $\in[0,1]$ for Wilcoxon rank-sum test on different faithfulness metrics. A larger $\rho$-statistic indicates a larger effect size. LAS: Label Adjusted Simulation, LRA: Label Rationale Agreement, and FTC: Faithfulness-by-Counterfactual (ours), and FTC-$\mathcal{W}$ and FTC-$\mathcal{K}$ ($\alpha=0.7$) are the Wasserstein and Kl-divergence variants. * indicates significance (p-value$<0.05$). The test is conducted between the two groups; data points where the human agrees vs disagrees with the model's prediction on counterfactual hypothesis. }
\label{tab:ranksum_uf}
\end{table}

In the following randomly sampled examples, the original Contradiction ($\texttt{C}$) label should result in Entailment ($\texttt{E}$) for the new counterfactual hypothesis. However the incompleteness or poor explanations result in the wrong label. 

\begin{figure}[!ht]

\begin{minipage}{0.5\textwidth}
\begin{center}
\includegraphics[width=5.5cm,height=4cm]{images/1876536922.jpg}
\caption{\small Example of low quality explanation where the counterfactual hypothesis does not result in the expected logical label.}
\vspace{0.5em}
\end{center}
\begin{minipage}{0.45\textwidth}
\small
\begin{tabular}{lp{16em}}
Hypothesis & Two men shooting each other. \\
Explanation & Dogs and men are different species. \\
Counterfactual &  Two dogs are shooting each other.  \\
Old$\rightarrow$New Label & Contradiction $\rightarrow$ Contradiction 
\end{tabular}
\end{minipage}

\end{minipage} \hfill

\begin{minipage}{0.5\textwidth}
\vspace{1em}
\begin{center}
\includegraphics[width=5.5cm,height=4cm]{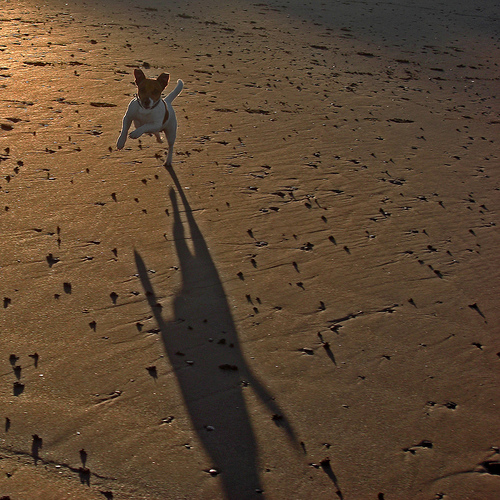}
\caption{\small Example of low quality explanation where the counterfactual hypothesis does not result in the expected logical label.}
\vspace{0.5em}
\end{center}
\begin{minipage}{0.45\textwidth}
\small
\begin{tabular}{lp{16em}}
Hypothesis & A dog is sitting on a porch. \\
Explanation &  A dog cannot be running and sitting at the same time. \\
Counterfactual &  A dog is running on a porch. \\
Old$\rightarrow$New Label & Contradiction $\rightarrow$ Contradiction 
\end{tabular}
\end{minipage}

\end{minipage} \hfill

\begin{minipage}{0.5\textwidth}
\vspace{1em}
\begin{center}
\includegraphics[width=5.5cm,height=4cm]{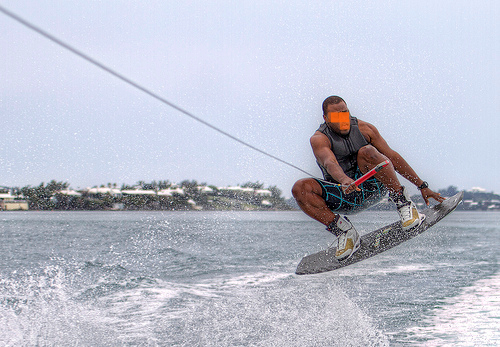}
\caption{\small Example of low quality explanation where the counterfactual hypothesis does not result in the expected logical label.}
\vspace{0.5em}
\end{center}
\begin{minipage}{0.5\textwidth}
\small
\begin{tabular}{lp{18em}}
Hypothesis & Man riding a motorcycle. \\
Explanation &  The man is on a lake with a monoboard. \\
Counterfactual &  Man riding a monoboard. \\
Old$\rightarrow$New Label & Contradiction $\rightarrow$ Contradiction 
\end{tabular}
\end{minipage}

\end{minipage} \hfill
\end{figure}

\onecolumn
\subsection{Hypothesis Rewriting Validation}

\begin{table*}[!h]
\centering
\resizebox{\columnwidth}{!}{%

\small
\begin{tabular}{ll@{\hskip 0.3in}cccc@{\hskip 0.3in}cccc}
\toprule
  &         &        \multicolumn{4}{c}{Human Explanations $z^* \mapsto x^{\textrm{cf}}$ } & \multicolumn{4}{c}{Generated Explanations $\hat{z} \mapsto x^{\textrm{cf}}$ } \\
Extract & Transform & \texttt{C} & \texttt{E} & $\texttt{N}_{[A\backslash B]}$ & $\texttt{N}_{[B\backslash A]}$ &  \texttt{C} & \texttt{E} & $\texttt{N}_{[A\backslash B]}$ & $\texttt{N}_{[B\backslash A]}$ \\
\midrule
regex & regex &  0.690 & 0.638 & 0.010 & 0.012 & 0.744 & 0.701 & 0.473 & 0.418 \\ 
\addlinespace
 regex & GPT2-xl & 0.696 & 0.690 & 0.716 & 0.748 & 0.813 & 0.817 & 0.795 & 0.727 \\
GPT2-xl & regex & 0.779 & 0.632 & 0.566 & 0.447 & 0.860 & 0.759 & 0.700 & 0.512 \\
GPT2-xl & GPT2-xl & 0.766 & 0.716 & 0.592 & 0.643 & 0.813 & 0.810 & 0.743 & 0.725 \\
\addlinespace
regex & Neo1.3B & 0.684 & 0.675 & 0.734 & 0.723 & 0.798 & 0.834 & \textbf{0.806} & 0.702 \\ 
Neo1.3B & regex & 0.774 & 0.643 & 0.473 & 0.414 & 0.886 & 0.787 & 0.668 & 0.565 \\ 
Neo1.3B & Neo1.3B & 0.769 & 0.698 & 0.716 & 0.721 & 0.809 & 0.822 & 0.798 & 0.703 \\ 
\addlinespace
regex & Neo2.7B & 0.690 & 0.709 & 0.710 & \textbf{ 0.752} & 0.840 & 0.860 & \textbf{0.805} & \textbf{0.714} \\
Neo2.7B & regex & 0.801 & 0.691 & 0.509 & 0.565 & \textbf{0.893} & 0.781 & 0.669 & 0.584 \\
Neo2.7B & Neo2.7B & \textbf{0.822} & \textbf{0.782} & \textbf{0.743} & \textbf{0.754} & 0.870 & \textbf{0.881} & \textbf{0.807} & 0.709 \\
\bottomrule
\end{tabular}}

\caption{\small METEOR scores for Hypothesis Revision by either using regex, GPT-Neo, or a combination of both. Bold is applied column-wise. We show the breakdown of revised hypothesis by class label (Contradiction \texttt{C}, Entailment \texttt{E} and Neutral \texttt{N}), and also whether the explanations are human or model generated. Note that there are two counterfactual hypothesis generated for \texttt{N} ($\texttt{N}_{[A\backslash B]}$ and $\texttt{N}_{[B\backslash A]}$) as described in \autoref{sec:satisfiability}: Proposition 3.  As expected, the performance of Neo2.7B $>$ Neo1.3B $>$ regex for both $z^*$ and $\hat{z}$ when the same `model' is used for both extract and transform. Considering a combination of approaches (row-wise), for $\hat{z}$, the best combination of (regex $+$ Neo1.3B) performs close to the best combination of (regex $+$ Neo2.7B), which suggests that a smaller $\mathcal{H}^{\textrm{model}}$ may be sufficiently competitive. }
\label{tab:validate_hyp_appendix}

\end{table*}

\subsection{Sensitivity Analysis}

\begin{table*}[!h]
\resizebox{\columnwidth}{!}{
\begin{tabular}{lll|rrrr|rrrr|rrrr}
&&& \multicolumn{4}{c}{FTC-$\mathcal{W}$} & \multicolumn{4}{c}{FTC-$\delta$} & \multicolumn{4}{c}{FTC-$\mathcal{K}$} \\
$x$ & $u$ & $y$ & gpt-gpt & gpt-regex & regex-gpt & regex-regex &      gpt-gpt & gpt-regex & regex-gpt & regex-regex & gpt-gpt & gpt-regex & regex-gpt & regex-regex \\

\midrule
 \checkmark  & \checkmark  & \checkmark     &   0.608 &          0.585 &          0.589 &                 0.571 &        0.684 &          0.585 &          0.656 &                 0.562 &   0.126 &         -0.112 &          0.032 &                -0.224 \\
\checkmark  & {}  & {}        &   0.539 &          0.534 &          0.536 &                 0.394 &        0.591 &          0.522 &          0.584 &                 0.281 &  -0.200 &         -0.299 &         -0.229 &                -1.407  \\ 
{}  & \checkmark  & {}        &   0.580 &          0.421 &          0.557 &                 0.423 &        0.650 &          0.340 &          0.618 &                 0.323 &  -0.055 &         -1.261 &         -0.182 &                -1.229 \\
{}  & {} & \checkmark        &   0.553 &          0.445 &          0.551 &                 0.447 &        0.621 &          0.396 &          0.613 &                 0.375 &  -0.245 &         -1.111 &         -0.225 &                -1.080 \\

\bottomrule
\end{tabular}}
\caption{Raw scores for FTC metric variants by perturbing inputs to the explanation generator (sensitivity analysis). $x$, $u$, and $y$ refers to hypothesis, premise, and label respectively. We vary different parts of the extract-transform pipeline (either with regex-regex, regex-gpt, gpt-regex, gpt-gpt), using GPT-Neo2.7B. We find that all columns have the first row as the largest number as expected, but FTC-$\mathcal{K}$ produces larger difference between the first row and other rows.}
\end{table*}

\clearpage

\subsection{Prompt templates}
\label{sec:prompt_templates}

\begin{table*}[!h]
\centering
\tiny
\resizebox{\columnwidth}{!}{
\begin{tabular}{p{7cm}p{3.5cm}p{3.5cm}}
\toprule
                                                                                                                                                                                                Explanation        &                                                 A &                                            B \\
\midrule
                                                                                                                                                                                       \texttt{Contradiction} &                                                   &                                              \\
                                                                                                                                                                                       
                                                                                                                                                           Jail cells aren't pastel-colored or a classroom.        &                                        jail cells &                pastel-colored or a classroom \\
                                                                                                         The two men cannot both be at a construction site and also never have been at a construction site.        &                            at a construction site &      never have been at a construction site. \\
                                                                                                                                                            One cannot be happy and angry at the same time.        &                                             happy &                                        angry \\
                                                                                                                                                                                        a dog is not a cat.        &                                               dog &                                          cat \\
                                                                                                                                                                                        Star is not balloon        &                                              star &                                      balloon \\
                                                                                                                                                                       \midrule                   \texttt{Entailment} &                                                   &                                              \\
                                                                                                                                                                 The guy on the phone is sitting at a desk.        &                                  guy on the phone &                            sitting at a desk \\
                                                                                                                                                                 People is a generalization of many people.        &                                            people &                                  many people \\
                                                                                                                                     Jockeys are people, so jockeys riding horses are people riding horses.        &                                           jockeys &                                       people \\
If people are sitting on blankets and the blankets are in the park, then the people are sitting at the park as well, because they cannot be in a separate location from the blankets they are sitting upon.        &                        people sitting on blankets &       people sitting on blankets in the park \\
                                                                                                                                         If your pausing for a photo then they are having their photo taken        &                               pausing for a photo &                           having photo taken \\
                                                                                                                                                               \midrule                              \texttt{Neutral} &                                                   &                                              \\
                                                                                                                Looking on does not mean waiting for their turn and doesn't mean they are at a competition.        &                                           looking &                       waiting for their turn and they are at a competition \\
                                                                                   Just because a track athlete is carrying a large pole it does not mean they gets ready for their turn at the pole vault.        &          A track athlete is carrying a large pole & A track athlete gets ready for their turn at the pole vault. \\
                                                                                                                        There is no evidence that the women talking to the man outside the building are sad & The  women talking to the man outside the building &                            the women are sad \\
                                                                                                                                                   Just because he has a mop, does not mean he has a broom.        &                                      he has a map &                               he has a broom \\
                                                                                                                                         No way to know that is alone because he lost his friends in a war.        &                                       he is alone &                 he lost his friends in a war \\
\bottomrule
\end{tabular}}
\caption{Example of prompts for each label, Contradiction, Entailment and Neutral provided to a few-shot priming model (GPT-Neo) for Extraction. We show the model 20 prompts in all experiments.}
\end{table*}

\begin{table*}[!h]
\centering
\tiny
\resizebox{\columnwidth}{!}{
\begin{tabular}{p{4.5cm}p{5cm}p{2cm}p{2cm}}
\toprule
 hyp\_original &                                                                                                                                                                       hyp\_cf &                                                 A &                                           B \\
\midrule

                                                                                        \texttt{Contradiction / Entailment} &                                                                                                                                                                              &                                                   &                                             \\
                                             They are waiting for parole in their jail cells. &                                                                                                                They are waiting for parole in their pastel-colored classroom &                                        jail cells &                    pastel-colored classroom \\

                                          The two men have never been to a construction site. &                                                                                                                                      The two men are at a construction site. &                            at a construction site &      never been to a construction site \\

                                                an angry man grills vegetables on a barbecue. &                                                                                                                                 A happy man grills vegetables on a barbecue. &                                             angry &                                       happy \\

                                                                       There is a cat running &                                                                                                                                                       There is a dog running &                                               cat &                                         dog \\

                        The young man is holding the balloon inside the large stone building. &                                                                                                           The young man is holding the star inside the large stone building. &                                           balloon &                                        star \\

                                                                                \midrule         \texttt{Neutral} &                                                                                                                                                                              &                                                   &                                             \\

                               The skateboarders are waiting for their turn at a competition. &                                                                                            \makecell[l]{A:\hspace{1em}The skateboarders are looking.; \\B:\hspace{1em}The skateboarders are waiting for their turn.} &                                           looking &                      waiting for their turn \\

                                 A track athlete gets ready for their turn at the pole vault. &                                                                     \makecell[l]{A:\hspace{1em}A track athlete carrying a large pole.; \\B:\hspace{1em}A track athlete gets ready for their turn at the pole vault.} &                             carrying a large pole & gets ready for their turn at the pole vault \\

                                Three sad women standing outside a building talking to a man. &                                                                                                           \makecell[l]{A:\hspace{1em}Three talking to the man outside the building; \\B:\hspace{1em}Three sad women} & the women talking to the man outside the building &                           the women are sad \\


                      There is an old man sitting alone because he lost his friends in a war. &                                                                                   \makecell[l]{A:\hspace{1em}There is an old man sitting alone; \\B:\hspace{1em}There is an old man who lost his friends in  a war.} &                                       he is alone &                he lost his friends in a war \\

\bottomrule
\end{tabular}}
\caption{Example of prompts for each label, Contradiction, Entailment and Neutral provided to a few-shot priming model (GPT-Neo) for Transformation. Contradiction and Entailment uses the same set of prompts. We show the model 20 prompts in all experiments.}
\end{table*}
\clearpage
\twocolumn
\subsection{Regex templates}

\begin{table}[!h]
\centering
\small
\begin{tabular}{l|p{4cm}}

R1& a/an, a type of, a way of saying, the same as, a rephrasing of, a/another form of, synonymous with \\
R2& is, are \\ 
R3& synonyms \\ 
R4& then, so, must be, has to be, have to be \\ 
\midrule
P1& A is R1 B \\ 
P2& A implies B \\ 
P3& A and B are R3 \\ 
P4& A and B R2 the same thing \\ 
P5& if A then B \\ 
P6& A R4 B \\ 
P7& A R2 B   \\ 

\end{tabular}

\caption{Regex for extracting logical variables ``A'' and ``B'' from Entailment (\texttt{E}).}
\label{tab:regexe}
\end{table}

\begin{table}[!h]
\centering
\small
\begin{tabular}{l|p{5cm}}

R1& cant, cannot, can't, can not \\
R2& at the same time,simultaneously, at once\\
R3& is, are\\
R4& not the same as, not, the opposite of, different than\\
R5& he, she, they\\
R6& a, an\\
\midrule
P1& A R3 not R6 B\\
P2& R1 be A and B R2\\
P3& A R1 be B\\
P4& A R3 R4 B\\
P5& R3 either A or B\\
P6& A R3 not B\\
P7& A R3 different than B\\
P8& R1 be A if R3 B\\
P9& R1 be A if R5 is B\\
P13& R1 A if B\\
P10& A and B R3 different\\
P11& A would not be able to B\\
P12& A R1 be B\\
    \end{tabular}
    \caption{Regex for extracting logical variables ``A'' and ``B'' from Contradiction (\texttt{C}).}
    \label{tab:regexc}
\end{table}

\begin{table}[!t]
\small
\begin{tabular}{l|p{5cm}}

R1& is, are \\
R2& not all, not every \\
R3& mean, necessarily mean, make, necessarily make, imply, indicate \\
R4& does not, doesnt, doesn't \\
R5& did not, didn't, didnt \\ 
\midrule
P1& R2 A R1 B \\ 
P2& there is more A than B \\ 
P3& there is more A than B \\ 
P4& just because A R4 R3 B \\ 
P5& A R1 not necessarily B \\
P6& A R4 have to be B \\ 
P7& A R4 necessarily B \\
P8& A R4 R3 B \\ 
P9& can A without B \\
P10& could be A not just B\\
P12& we R5 know A to B\\
P11& we R5 know if A or B\\
P12& we can't tell if A is B\\
P13& if A then B\\
P14& this R4 imply A or B\\
P15& A and B R1 two different\\
P16& A and B R1 different\\
P17& not everyone A will B\\
P18& A may not be B\\
P19& it cannot be assumed that A is B\\
P20& some A or B\\
P21& A might not be B\\
P22& there is not evidence A or B\\
P23& R4 have to be A to B\\
P24& no way to know A or B \\         

\end{tabular}

\caption{Regex for extracting logical variables ``A'' and ``B'' from Neutral (\texttt{N}).}
\label{tab:regexn}
\end{table}

\end{document}